\documentclass[sigconf]{acmart}

\usepackage{algorithm}
\usepackage{algpseudocode}
\usepackage{microtype}
\usepackage{booktabs}
\usepackage{graphicx}
\usepackage{multirow}
\usepackage{multicol}
\usepackage{diagbox}
\usepackage{enumitem}

\AtBeginDocument{%
  }

\copyrightyear{2024}
\acmYear{2024}
\setcopyright{rightsretained}
\acmConference[MCHM '24]{Proceedings of the 1st International Workshop on Multimedia Computing for Health and Medicine}{October 28-November 1, 2024}{Melbourne, VIC, Australia}
\acmBooktitle{Proceedings of the 1st International Workshop on Multimedia Computing for Health and Medicine (MCHM '24), October 28-November 1, 2024, Melbourne, VIC, Australia}
\acmDOI{10.1145/3688868.3689193}
\acmISBN{979-8-4007-1195-4/24/10}

\begin{document}

\title{Revisiting Surgical Instrument Segmentation Without Human Intervention: A Graph Partitioning View}

%


\author{Mingyu Sheng}
\orcid{0000-0003-3567-0883}
\affiliation{%
  \institution{The University of Sydney}
  \city{Sydney}
  \state{NSW}
  \country{Australia}
}
\email{mshe0136@uni.sydney.edu.au}

\author{Jianan Fan}
\orcid{0000-0003-4424-9572}
\affiliation{%
  \institution{The University of Sydney}
  \city{Sydney}
  \state{NSW}
  \country{Australia}
}
\email{jfan6480@uni.sydney.edu.au}

\author{Dongnan Liu}
\orcid{0000-0001-8102-3949}
\affiliation{%
  \institution{The University of Sydney}
  \city{Sydney}
  \state{NSW}
  \country{Australia}
}
\email{dongnan.liu@sydney.edu.au}

\author{Ron Kikinis}
\orcid{0000-0001-7227-7058}
\affiliation{%
  \institution{Harvard Medical School}
  \city{Boston}
  \state{MA}
  \country{USA}
}
\email{kikinis@bwh.harvard.edu}

\author{Weidong Cai}
\orcid{0000-0003-3706-8896}
\affiliation{%
  \institution{The University of Sydney}
  \city{Sydney}
  \state{NSW}
  \country{Australia}
}
\email{tom.cai@sydney.edu.au}





\begin{abstract}
    Surgical instrument segmentation (SIS) on endoscopic images stands as a long-standing and essential task in the context of computer-assisted interventions for boosting minimally invasive surgery. 
    Given the recent surge of deep learning methodologies and their data-hungry nature, training a neural predictive model based on massive expert-curated annotations has been dominating and served as an off-the-shelf approach in the field, which could, however, impose prohibitive burden to clinicians for preparing fine-grained pixel-wise labels corresponding to the collected surgical video frames.
    In this work, we propose an unsupervised method by reframing the video frame segmentation as a graph partitioning problem and regarding image pixels as graph nodes, which is significantly different from the previous efforts. 
    A self-supervised pre-trained model is firstly leveraged as a feature extractor to capture high-level semantic features. 
    Then, Laplacian matrixs are computed from the features and are eigendecomposed for graph partitioning. 
    On the "deep" eigenvectors, a surgical video frame is meaningfully segmented into different modules such as tools and tissues, providing distinguishable semantic information like locations, classes, and relations. 
    The segmentation problem can then be naturally tackled by applying clustering or threshold on the eigenvectors.
    Extensive experiments are conducted on various datasets (e.g., EndoVis2017, EndoVis2018, UCL, etc.) for different clinical endpoints. 
    Across all the challenging scenarios, our method demonstrates outstanding performance and robustness higher than unsupervised state-of-the-art (SOTA) methods. The code is released at \href{https://github.com/MingyuShengSMY/GraphClusteringSIS.git}{https://github.com/MingyuShengSMY/GraphClusteringSIS.git}.
    
    

\end{abstract}




\maketitle

\section{Introduction}  
\label{Introduction}
    Minimally invasive surgery (MIS) offers several advantages over standard open surgery, such as reduced pain, lower risk, and shorter recovery period \cite{maier2017surgical}. Its extensive use of endoscopic cameras for probing human body allows surgeons to observe pathological tissues and manage surgical tools effectively. Despite the advantages of MIS, this technique still faces various challenges, including long surgical procedures, intricate tool operations, limited fields of view, and challenging hand-eye coordination \cite{rueckert2024corrigendum}.

    Interest in addressing the limitations of MIS has grown significantly in recent years when various techniques have been developed to help surgeons overcome challenges, such as professional training and evaluation, procedure analysis and optimization, and visual-based frame processing \cite{haidegger2022robot, Yue10098668Cascade}. 
    One effective approach is to utilize positional and movement data of surgical instruments during operations. It can be accomplished with infrared and electromagnetic tracking or external markers directly attached to the tools \cite{bouarfa2012vivo}. However, these methods require additional efforts and delicate preparations, which renders it troublesome to integrate those techniques into existing surgical workflows \cite{wang2022visual}.
    
    Due to these challenges, recent efforts have focused on visual-based endoscopic frame processing. 
    With the explosive growth of machine learning techniques and the promoted accessibility of computational resources, a stream of visual-based approaches has been proposed. 
    Hand-crafted-feature-based object detection appears as a classic method to localize surgical instruments and track them in a video stream \cite{qiu2019real}. 
    While those approaches facilitate high-speed frame processing, the detection results are often unsatisfactory due to the varying positions, orientations, and shapes of surgical instruments. 
    In an endoscopic video frame, surgical tools typically appear from the corners or edges towards the center, making traditional axis-aligned bounding box detection methods suboptimal \cite{rueckert2024corrigendum}. 
    In this context, surgical instrument segmentation (SIS) enables more accurate predictions of surgical instruments at the pixel level and therefore stands as a promising pathway for providing vital assistance to surgeons \cite{Yue_Zhang_Hu_Xia_Luo_Wang_2024}.

    
    

    However, tremendous labeled ground-truth data are required at the price of training a high-performance supervised SIS model, and several obstacles could hinder the data acquisition procedure. 
    Obtaining and collecting labeled data would rely on the deep involvement of domain experts to distinguish various specific surgical tools, pathological tissues, and organs, which could consume massive labour efforts. 
    Accessing surgical image data might be further restrained due to patient privacy issues. 
    In this regard, semi-supervised and unsupervised methods for SIS arise as worthwhile topics to explore. Compared to semi-supervised learning, unsupervised learning is fully annotation-free and is capable of prospecting more general and high-level patterns (e.g., semantic correlation) hiding in data distribution \cite{liu2021graph, zhao2021anchor, da2019self}. 


    Given its intriguing potential, several unsupervised SIS models have been developed \cite{sestini2023fun, zhao2021anchor}, where the pseudo-label technique is widely used and plays an important role. 
    The pseudo-label approach provides a straightforward solution to convert an unsupervised learning task into a pseudo-supervised one, thereby being easily implemented for training neural networks. 
    Low-level features are the majority materials to derive pseudo-labels, such as color thresholds and edge detection \cite{van2021unsupervised}. 
    Alternatively, pseudo-labels could also be generated by clustering the deep feature obtained from an encoder model \cite{larsson2019fine}.
    
    On the other hand, the effectiveness of the above technique highly relies on the quality and accuracy of the pseudo-label. 
    Therefore, it may suffer from certain limitations in the SIS field, especially when the pseudo-labels are generated from low-level features. For example, most endoscopic images are likely of low quality, containing significant noise and blur; the scenes of endoscopic images involve intermixed tissues with irregular shapes, colors, and fuzzy tissue connectives. As a result, the quality of pseudo-labels generated from low-level features may be compromised.

    In a nutshell, the research gaps in the current field of SIS are summarized as:
	1) the performance and robustness of supervised models are limited due to the scarcity of labeled data; 2) the pseudo-label technique may lead to erratic or unstable outcomes, especially with low-quality and intricate endoscopic images; and 3) there is a notable lack of studies on label-free methods for the SIS task.
 
    
    To overcome the above problems, we devise a label-free unsupervised method. A self-supervised pre-trained model is leveraged as a feature extractor, based on Vision Transformer (ViT) to capture the global high-level context features from surgical video frames. We map the segmentation task into a graph-partitioning problem by eigendecomposing a "deep" Laplacian matrix calculated from the deep dense features. The segmentation mask can be predicted by conducting clustering or thresholding on top of the eigenvectors.
    Our method outperforms unsupervised SOTA methods on both binary and multi-class segmentation tasks and demonstrates significant robustness across different datasets, as evidenced by the experimental results.

    Our key contributions are summarized as below: 
    \setlist{nolistsep}
    \begin{itemize}
        \item We propose an unsupervised method that can handle all categories of surgical instrument segmentation (SIS) tasks, including binary, part, type, and semantic segmentation tasks.

        \item We develop a novel framework that solves the SIS task via graph theory in a graph-cutting manner, that the high-level context features are regarded as a graph and its pixels as nodes.

        \item We introduce unsupervised graph clustering applied on eigenvectors decomposed from the Laplacian matrix of deep features, and unsupervised salient detection based on the Fiedler Vector, where the most apparent object is effectively detected in surgical video frames.

        \item We test our method on various datasets, conducting extensive and comprehensive experiments, to demonstrate our method's SOTA performance and robustness.
    \end{itemize}
    

    In Section \ref{RelatedWork}, we reviewed related studies, including recent unsupervised segmentation methods in both the natural image field and the SIS field. Our approach is then introduced in Section \ref{Method}. The experiment details and results are reported and analyzed in Section \ref{ExperimentDiscussion}, followed by the conclusion and future expectations in Section \ref{Conclusion}.

\section{Related Work}
\label{RelatedWork}

    \subsection{Unsupervised Image Segmentation} 
    \label{RelatedWork: UnsupervisedImageSegmentation}
        Compared with image classification, image segmentation is a pixel-wise classification task. There are three main categories for image segmentation: salient segmentation, semantic segmentation, and instance segmentation. Salient segmentation (also called salient detection) aims to segment the most salient object or a specific object as foreground against the background. 
        Semantic segmentation is a multi-class task that segments and classifies all labeled objects according to their semantic features. Instance segmentation is also a multi-class task, but every object is considered a unique instance or class. Unsupervised segmentation indicates that no annotated ground-truth is provided for training.

        For the earliest works before 2018, unsupervised image segmentation was typically addressed by using traditional machine learning and computer vision methods, such as threshold, watershed method, and Markov Random Field (MRF) \cite{chang2011co, zhou2011n, haller2017unsupervised}. Nowadays, deep learning techniques are widely leveraged in this field \cite{Fan10378316Taxonomy, ke2022unsupervised, ma2021self, hamilton2022unsupervised, Liu9156759Unsupervised}. In our work, unsupervised image segmentation methods are basically classified into two types: \textit{Pseudo-Label} based and \textit{Label-Free} based approaches.
        
        \noindent
        \textbf{Pseudo-Label}. An alternative approach for unsupervised image segmentation is using pseudo-label as supervision. The pseudo-label is usually generated from low-level features, predicted from pre-trained models, or clustered from the output feature vectors \cite{larsson2019fine, ke2022unsupervised}, such as PiCIE \cite{cho2021picie} via clustering and MaskContrast \cite{van2021unsupervised} via a pre-trained model. When the pseudo-labels are generated from clustering, the generating and training processes proceed by loop until convergence. Regarding the pseudo-labels as hints and supervision, contrastive learning is a popular training strategy adopted in pseudo-label-based methods to increase and decrease the similarity between two features according to their correlation \cite{cho2021picie, van2021unsupervised, ke2022unsupervised}. 
        However, the pseudo-label technique can be unreliable and inaccurate for tasks containing complex scenes and low-quality images (e.g., blur, light reflection, and narrow perspective in endoscopic image frames), and thus uncertainly trains the model. Weak robustness is another problem, for which a model trained on a dataset with its pseudo-labels as supervision cannot be effortlessly generalized to other situations/scenes. 

        \noindent
        \textbf{Label-Free}. Extracting high-level dense features via an image encoder and then clustering (usually by K-Means) is a general framework for label-free methods. A promising approach to further enhance the model performance is by extending the encoder with a segmentation head and optimizing/training the head by maximizing the consistency among output feature maps extracted from an image and its random augmentations or nearest neighbors \cite{hung2019scops, ji2019invariant}, such as STEGO \cite{hamilton2022unsupervised}.
        \textit{Graph method} is recently a new topic for unsupervised learning because of its outstanding accuracy and robustness over different data distributions. It transforms image segmentation tasks into graph-partitioning tasks by treating pixels as nodes and their similarities as edges
        \cite{melas2022deep, wang2023cut, Fan_2024_Seeing}. In addition, another promising approach is directly generating segmentation masks by training a \textit{Generative Adversarial Network (GAN)}. To date, most GAN-based methods are proposed mainly focusing on salient segmentation, such as \cite{bielski2019emergence, chen2019unsupervised, benny2020onegan, abdal2021labels4free, ma2021self}. Nevertheless, how to efficiently and stably train a GAN model is an inevitable challenge that would be more severe due to insufficient data.

    
    

    \subsection{Surgical Instrument Segmentation} 
    \label{RelatedWork: SurgicalInstrumentSegmentation}
        Surgical instrument segmentation (SIS) is to classify every single pixel of a medical endoscopic image into a specific class including background, instrument, and pathological tissue. There are four categories of SIS tasks: \textit{Binary}, \textit{Part}, \textit{Type}, and \textit{Semantic Segmentation}. Binary segmentation is to segment the instrument and background; part segmentation extends it to classify different parts of instruments, like shaft, wrist, and clasper; type segmentation aims to identify different instrument types, like clamps, suturing needles, and threads; semantic segmentation categorizes all objects in an endoscopic image frame, including instruments, tissues, organs, and even bleeding areas.
        

        \noindent
        \textbf{Supervised SIS.} \cite{bouget2015detecting} firstly proposed a detector based on Support Vector Machine (SVM), designed for instrument segmentation and attitude estimation; their work marked a significant beginning in machine-learning-based surgical tool segmentation. Following their work, an increasing number of studies \cite{laina2017concurrent, attia2017surgical, garcia2017toolnet, garcia2017real} are proposed, leveraging deep learning techniques (e.g., LSTM, CNN, etc.). Various classic backbones were utilized in these works, such as FCN \cite{long2015fully}, U-Net \cite{ronneberger2015u}, and ResNet \cite{he2016deep}. To further improve performance, multi-scale spatial features were widely employed to capture more features including low-level and high-level semantic features \cite{jin2019incorporating, ni2019rasnet, islam2019real, ni2020attention, ni2022surginet, ceron2022real, shen2023branch}. In addition, attempts to extract temporal features from surgical videos have been made for fully leveraging the semantic correlation among continuous frames \cite{gonzalez2020isinet, wang2021efficient}. However, the performance and generalization ability of supervised methods are constrained by data scarcity.
        
        \noindent
        \textbf{Unsupervised SIS.} \cite{liu2020unsupervised} firstly proposed an unsupervised method called AGSD for the binary SIS task by utilizing pseudo-labels generated from low-level features (e.g., color and lightness). They assume that most surgical instruments are displayed with higher lightness and plainer colors than background tissues. This strategy may be effective for simple scenes but can be challenged in other more complex scenes, and cannot tackle multi-class segmentation. Similarly, \cite{chen2023surgnet} adopted Region Adjacency Graph (RAG) to generate pseudo-label and trained a model based on Masked Autoencoders (MAE) \cite{he2022masked}, whereas their main target is to segment blood vessels instead of instruments. \cite{sestini2023fun} leveraged optical flow for supervision and shape-prior as a hint to train a model based on the Teacher-Student structure. Nevertheless, the recent pseudo-label-based methods in the SIS field are restrained to binary segmentation, and robustness is limited due to the drawbacks and limitations of low-level pseudo-labels. In this study, recognizing the above limitations and inspired by the graph theory, we proposed a label-free method based on graph cutting, demonstrating the SOTA performance and robustness on various SIS datasets.
        

\section{Method}
\label{Method}
    \begin{figure*}[htbp]
    \centering
    
    \includegraphics[width=0.9\linewidth]{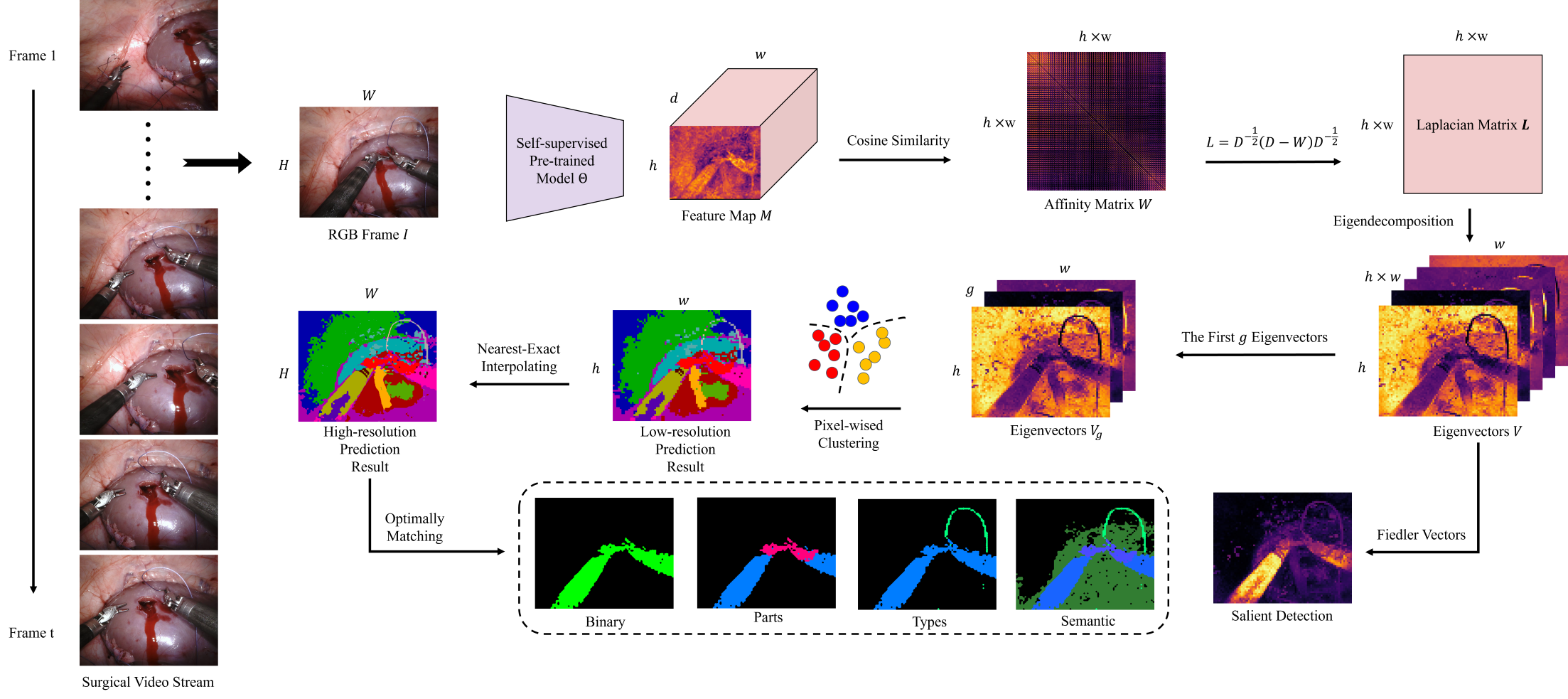}
    
    \caption{Overview of Our Method. \textmd{
    Every surgical video frame is fed into a ViT-based feature extractor to generate high-level dense features. Then, an affinity matrix $W$ is computed and its Laplacian matrix $L$ is calculated for the subsequent eigendecomposition, from which eigenvectors provide distinct features to distinguish different modules in a frame, where the first $g$ eigenvectors are stacked together for clustering and the second eigenvector (the Fiedler Vector) is leveraged for salient detection. 
    }
    }
    \label{Fig: overview}
    \Description{}
    \end{figure*}

    \begin{figure}[htbp] 
    \centering
    
    \includegraphics[width=0.8\linewidth]{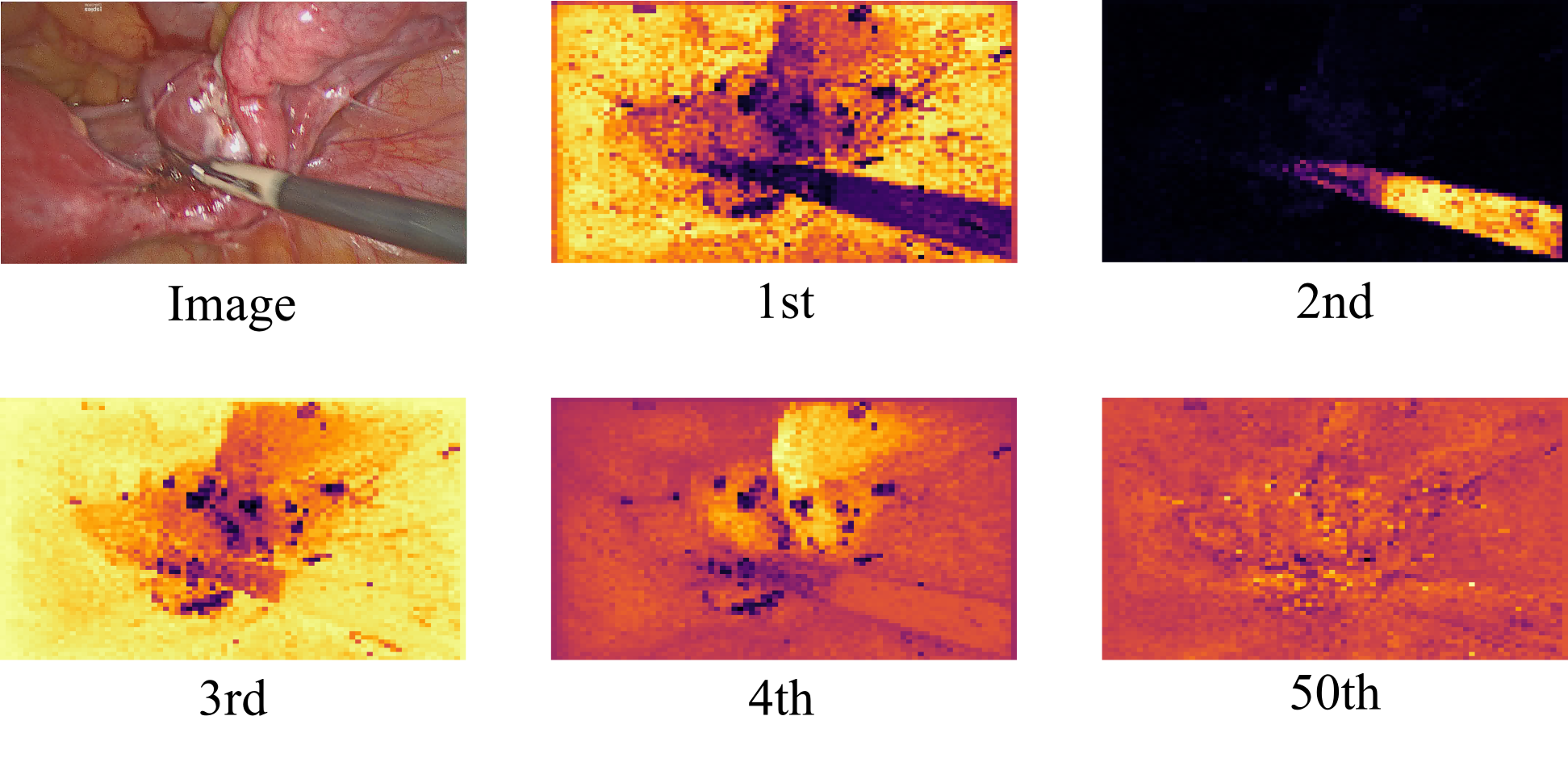}
    
    \caption{Sample Result of Eigendecomposition. \textmd{The top-left is the origin image frame, and "$i$ th" represents a visualized eigenvector with the $i$-th smallest eigenvalue.
    }
    }
    \label{Fig: eigenSample}
    \Description{}
    \end{figure}
    
    Our proposed approach is illustrated in Figure \ref{Fig: overview}. It takes a single frame from a surgical video stream as input whose deep context feature map is extracted from a self-supervised pre-trained model. Then, the Laplacian matrix is calculated from the feature map for the subsequent partitioning progress (i.e., graph clustering and salient detection).

    \subsection{Backbone}
    \label{Method: FeatureExtractor}
        We employ a self-supervised pre-trained model named DINO (developed by \cite{caron2021emerging} based on ViT) as a feature extractor, noted as $\Theta$. The DINO is trained in a distillation approach and can effectively capture global high-level context information. Let $I \in \mathbb{R}^{H \times W \times 3}$ do an RGB frame from a surgical video, where $H$ and $W$ are the frame height and width, respectively. Let $M \in \mathbb{R}^{h \times w \times d}$ represent the extracted feature map, where $d$ indicates feature channels, $h$ and $w$ are the height and width, respectively. The feature extraction is denoted by $M = \Theta(I)$. Then, the feature map $M$ is reshaped into a feature matrix $F \in \mathbb{R}^{s \times d}$ for the subsequent procedures, where $s = h \times w$ and its row $f_i \in \mathbb{R}^{d}$ represents a feature vector of pixel $i$.

    \subsection{Affinity Matrix Computation}
    \label{Method: AffinityMatrixComputation}
        Assuming the feature matrix $F$ as a graph $G = (P, E)$ where $P$ is the set of nodes/pixels, and $E$ is the set of edges/similarities. A positive semi-definite affinity matrix $W \in \mathbb{R}^{s \times s}$ is computed to represent the similarities among pixels, whose element is denoted by $w_(i, j)$ measuring the affinity between pixels $i$ and $j$. The computation of the affinity matrix $W$ follows: 
        
        \begin{equation}
        \label{Equ: SimMatrix}
          w_{i, j} = 
            \begin{cases}
              \text{cos}(f_i, f_j) & i \neq j \; \text{and} \; \text{cos}(f_i, f_j) > 0 \\
              0 & \text{otherwise}
            \end{cases},       
        \end{equation}
        where $\text{cos}(\cdot , \cdot)$ indicates cosine similarity between two vectors. Apart from normalizing the cosine similarity into [0, 1] to get the positive semi-definite affinity matrix, we particularly threshold the similarity by 0, because in our preliminary experiments, eigendecomposing a dense matrix is extremely time-consuming, and sometimes even fails due to ill-conditioning.
    
    \subsection{Eigendecomposition of Laplacian Matrix}
    \label{Method: EigenLapMatrix}
        \textbf{Background.} Normalized Cut \cite{Ncut} introduces that the eigenvectors decomposed from the Laplacian matrix display different modules/partitions in a graph, which conclusion is derived from minimizing a normalized graph cut cost denoted by:
        \begin{equation}
        \label{Equ: NCutCost}
            Ncut(A, B) = \frac{\sum_{i \in A, j \in B} w_{i, j}}{\sum_{i \in A, j \in P} w_{i, j}} + \frac{\sum_{i \in A, j \in B} w_{i, j}}{\sum_{i \in B, j \in P} w_{i, j}}.
        \end{equation}
        Eigenvectors with smaller eigenvalues indicate lower cut cost, thus presenting more accurate cutting for a graph. Given the matrix $W$, its Laplacian matrix is computed by: 
        
        \begin{equation}
        \label{Equ: NormLapM}
            L = D^{-1/2} (D - W) D^{-1/2},
        \end{equation}
        where $D \in \mathbb{R}^{s \times s}$ is a diagonal matrix whose entries are the row-summation of $W$. 
        
        In this study, we apply eigendecomposition on the "deep" Laplacian matrix. Let $V \in \mathbb{R}^{s \times s}$ note column-stacked eigenvectors sorted in ascent according to their eigenvalues, where the $i$-th smallest eigenvector is represented by $v_i \in \mathbb{R}^s$. The eigenvectors effectively express informative and distinctive modules in the surgical image. Some of them for a surgical video frame are presented in Figure \ref{Fig: eigenSample}, in which the eigenvectors with lower eigenvalues such as "1st" and "2nd" meaningfully segment the endoscopic image frame, where the background and surgical tool are feasibly distinguished. The eigenvectors with comparatively higher eigenvalues like "3rd" and "4th" also demonstrate informative modules in the image, where the background is further finely segmented into different semantic modules (e.g., light reflection and operated target tissues). However, as the eigenvalue increases, the eigenvector becomes chaotic and noise-like, such as "50th". With the eigenvectors, the two different strategies for graph cutting are introduced in the following sections.
    

    \subsection{Salient Detection}
    \label{Method: SalientDetection}
        The eigenvector with the second-smallest eigenvalue is called Fiedler Vector \cite{FiedlerEigen} denoted by $v_2$, significantly detecting the most salient object in an image. For example, the "2nd" in Figure \ref{Fig: eigenSample} effectively detects the surgical tool with high magnitudes on its shaft and head. In this case, a binary segmentation mask can be obtained by thresholding the Fiedler Vector with a certain value often with 0. We follow the common threshold method, setting the elements greater than 0 as foreground, and background otherwise. This binary-partitioning method can only deal with the binary segmentation task. Therefore, another more general method is introduced in the next section to deal with all four segmentation tasks. 

    \subsection{Graph Clustering}
    \label{Method: GraphClustering}
        In this section, the unsupervised segmentation task is solved via clustering. Firstly, informative eigenvectors are selected according to their eigenvalues. The first $g$ smallest eigenvectors are selected because lower eigenvalues indicate more meaningful feature representation possessed. Let $V_g \in \mathbb{R}^{s \times g}$ note the stacked selected eigenvectors. By reshaping the $V_g$ into $\hat{V}_g \in \mathbb{R}^{h \times w \times g}$, an embedded feature map is obtained, consisting of the meaningful eigenvectors. The K-Means algorithm with the given cluster number $k$ is implemented on the new feature map to cluster the pixels into different partitions. The clustering result is denoted by $\hat{Y} \in \mathbb{R}^{h \times w}$.

    \subsection{Interpolation}
    \label{Method: Interpolation}
        The preliminary prediction is a low-resolution label mask. The final high-resolution prediction noted by $Y \in \mathbb{R}^{H \times W}$ is calculated by Nearest-Exact Interpolation (NEI). Compared to the normal Nearest-Neighbor Interpolation (NNI), the NEI searches for a more accurate value by considering more pixels around the position instead of only one. For example, at a given position, the NNI only finds one pixel closest to the position and directly assigns the pixel value to it; in contrast, the NEI searches for the majority value around the interpolating position and provides precise interpolation.
        
    \subsection{Optimal Matching}
    \label{Method: OptimalMatching}
        \textbf{Our method is fully unsupervised without any access to the ground-truth labels.} Therefore, the prediction labels are optimally matched to the ground-truth labels for visualization and quantitative analysis. We employed two matching strategies: Hungarian matching and Majority-Vote matching. Hungarian matching is used to uniquely match each ground-truth label with every prediction if the numbers of clusters and classes are equal, otherwise, Majority-Vote matching is leveraged, which means a ground-truth label may be repeatedly matched to multiple prediction labels. 

\section{Experiment and Results
\label{ExperimentDiscussion}}

    \subsection{Evaluation Metric and Datasets}
    \label{ExpDis: Datasets}
        \textbf{Evaluation Metric}. The mean Intersection over Union (mIoU) is calculated to analyze the method performance. mIoU is an averaged IoU over all classes, in which IoU is a pixel-wised measure that comprehensively reflects the segmentation accuracy.
        
        \noindent
        \textbf{Datasets}. Our method is tested on five benchmark datasets for four segmentation tasks. 
        The complexity, difficulty, and object classes are various across the datasets, significantly challenging the method's performance and generalization ability. 
        \textit{EndoVis2017}, a dataset released in 2017 MICCAI EndoVis Robotic Instrument Segmentation Challenge \cite{allan20192017}. Binary, part, and type segmentation are available in this dataset. Similar to EndoVis2017, \textit{EndoVis2018} was released in 2018 \cite{allan20202018}. Binary, part, type, and semantic segmentation tasks are available for this dataset. \textit{ARTNetDataset} is an image-based dataset proposed and annotated by \cite{hasan2021detection}. Only binary labels are officially provided. \textit{CholecSeg8k} \cite{hong2020cholecseg8k} is a subset of the endoscopic dataset Cholec80 \cite{Cholec80}. Only the semantic segmentation is officially available in this dataset. \textit{UCL}. An ex-vivo synthetic dataset consists of 20 videos \cite{colleoni2020synthetic}. Scenes in UCL are distinctly divergent from the above in-vivo datasets. Binary ground-truth labels are officially provided. 

    \subsection{Implementation Details}
    \label{ExpDis: ImplementationDetails}
        The backbone model DINO is loaded as a feature extractor via Pytorch from \href{https://github.com/facebookresearch/dino}{https://github.com/facebookresearch/dino}. Following the existing works \cite{simeoni2021localizing, melas2022deep}, the deep features in the key vector at the last attention layer are extracted for computing the affinity matrix because of its better localization performance. The number of clusters is set as $k=15$. This is because we refer to the number of labeled classes in EndoVis2018 (12 classes) and CHolecSeg8k (13 classes), where the "background" is a class in the datasets, which can be further finely segmented into more classes (e.g., bleeding blood, tissue connectives, etc.), thereby we determine a slightly larger number for clusters. We set $g=k$ because each eigenvector represents a particular module in a frame, and we assume clustering with the number of modules can yield relatively superior performance, which is partially verified in our ablation study in Section \ref{ExpDis: AblationStudy}.

    \subsection{Comparison Results with SOTA Methods}
    \label{ExpDis: ResultsComparisonWithSOTAMethods}
        
        \noindent
        \textbf{Binary Segmentation Task.} Tables \ref{Tab: BinaryComparison} and \ref{Tab: BinaryComparisonGeneral} report comparisons between our method and SOTA methods, in two aspects of performance and robustness. The three kinds of supervision are noted as "Sup.", "Semi." and "Unsup." for supervised, semi-supervised and unsupervised methods respectively. The underline indicates the best results for all methods, and the best results for unsupervised methods are highlighted in bold. "SAL" and "CLU" represent "salient detection" and "graph clustering" corresponding to the Sections \ref{Method: SalientDetection} and \ref{Method: GraphClustering}. AGSD is reproduced in our study but cannot be trained on the non-video dataset (i.e., ARTNetDataset), so there is no corresponding result in Table \ref{Tab: BinaryComparison} for AGSD on ARTNetDataset.

        \begin{table}[htbp]
        \caption{
        Binary Segmentation Performance \textmd{(mIoU [\%])}. 
        \textmd{Methods noted with "$*$" are reproduced. "CLU" indicate our method based on graph clustering. \textbf{Bold} marks the best result across unsupervised methods. \underline{Underline} marks the best result across all methods.}
        }

        \label{Tab: BinaryComparison}
        \centering
        \resizebox{1.0\linewidth}{!}{
            \begin{tabular}{l c c c c c}
        
                \toprule 
                ~ & Supervision & EndoVis2017 & EndoVis2018 & ARTNetDataset \\ 
                \midrule 
                \midrule 
                ART-Net \cite{hasan2021detection} & \multirow{4}{*}{Sup.} & 81.00 & - & \underline{88.20} \\
                DRLIS \cite{pakhomov2019deep} & ~ & \underline{89.60} & - & -\\
                MF-TAPNet \cite{jin2019incorporating} & ~ & 87.56 & - & - \\ 
                \midrule 
                AOMA \cite{zhao2021anchor} & \multirow{2}{*}{Semi.} & 77.10 & 68.40 & - \\
                Duel-MF \cite{zhao2020learning} & ~ & 84.05 & - & - \\
                \midrule 
                FUN-SIS \cite{sestini2023fun} & \multirow{3}{*}{Unsup.} & 76.25 & - & - \\ 
                AGSD* \cite{liu2020unsupervised} & ~ & 81.08 & 63.83 & - \\ 
                Ours (CLU) & ~ & \textbf{81.12} & \textbf{\underline{79.46}} & 84.58 \\ 
                \bottomrule 
            \end{tabular}
            }
        \end{table}
        
        On the EndoVis2017 dataset, Table \ref{Tab: BinaryComparison} illustrates that our "CLU" method slightly outperforms the AGSD, and distinctly outperforms FUN-SIS by about 5\%. Regarding the EndoVis2018 dataset, our method shows dramatically high performance at 79.46\% higher than the AGSD by about 17\%. The severe performance degeneration for AGSD from 81.08\% on EndoVis2017 to 63.83\% on EndoVis2018 is because the AGSD method is a pseudo-label-based method trained with pseudo-labels generated from low-level features such as color and lightness, whose effectiveness is severely deducted due to the complex surgical scene in the EndoVis2018 dataset. Compared with some supervised and semi-supervised methods like ART-Net and AOMA, our method demonstrates higher performance, but slightly lower than most supervised methods like Duel-MF and DRLIS.

        \begin{table}[t]
        \caption{
        Binary Segmentation Robustness \textmd{(mIoU [\%])}.
        \textmd{"SAL" indicates our method based on salient detection.}
        }
        \label{Tab: BinaryComparisonGeneral}
        \centering
        \resizebox{1.0\linewidth}{!}{
            \begin{tabular}{l c c c c c}
        
                \toprule 
                ~ & EndoVis2017 & EndoVis2018 & UCL & Avg. & Std.\\ 
                \midrule
                \midrule
                AGSD* \cite{liu2020unsupervised} & 81.08 & 63.83 & 78.94 & 74.62 & $\pm$ 7.68 \\ 
                AGSD* (o-1) & 72.52 & 78.70 & 58.71 & 69.98 & $\pm$ 8.36 \\ 
                Ours (SAL) & 65.09 & 64.04 & 65.51 & 64.88 & \textbf{$\pm$ 0.62} \\ 
                Ours (CLU) & \textbf{81.12} & \textbf{79.46} & \textbf{79.75} & \textbf{80.11} & $\pm$ 0.72 \\ 
                \bottomrule 
                
            \end{tabular}
            }
        \end{table}
        
        For the generalization ability comparison across unsupervised methods, as reported in Table \ref{Tab: BinaryComparisonGeneral}, the SOTA method AGSD shows comparatively weaker robustness indicated by the high standard deviation of $\pm$ 7.68\% across the three datasets, higher than ours by 7.06\% and 6.96\% for "SAL" ($\pm$ 0.62\%) and "CLU" ($\pm$ 0.72\%) respectively. Performance degeneration of AGSD is exposed when testing on unseen datasets, except for the EndoVis2018 dataset, where mIoU increases, because the quality and reliability of pseudo-labels for EndoVis2018 are severely disrupted due to its high complexity, thus the "AGSD" is under-fitted on EndoVis2018 compared to "AGSD (o-1)". On the other hand, when AGSD is trained on the two EndoVis datasets and tested on the UCL dataset, the mIoU severely declines to 58.71\% due to the distinctly different scenes in the UCL dataset, such as tissue textures, lightness, and angles of view. In contrast, our method demonstrates outstanding robustness across different datasets justified with a small standard deviation at only $\pm$ 0.62\% and $\pm$ 0.72\% for "SAL" and "CLU" respectively. In particular, our "SAL" method shows inferior performance than others because only one eigenvector (the Fiedler Vector) is used, and its dramatic stability across the three datasets indicates its capability for stable detection on a wider range of scenes. 


        \noindent
        \textbf{Multi-class Segmentation Tasks.} The recent SOTA unsupervised methods cannot tackle multi-class segmentation tasks (i.e., part, type, and semantics) because of the limitations of pseudo-labels that can only reflect pixel labels in binary. In contrast, our CLU method is capable of both binary and multi-class segmentation tasks. We report the experimental results in Tables \ref{Tab: PartsComparison}, \ref{Tab: TypesComparison}, and \ref{Tab: SemanticComparison} for part, type, and semantic segmentation respectively.  

        \begin{table}[htbp]
        \caption{
        Part Segmentation Performance \textmd{(mIoU [\%])}.
        }
        \label{Tab: PartsComparison}
        \centering
        \resizebox{1.0\linewidth}{!}{
            \begin{tabular}{l c c c c c}
        
                \toprule 
                ~ & Supervision & EndoVis2017 & EndoVis2018 \\ 
                \midrule 
                \midrule 
                DRLIS \cite{pakhomov2019deep} & \multirow{3}{*}{Sup.} & \underline{76.40} & - \\
                DMNet \cite{wang2021efficient} & ~ & - & \underline{67.50} \\
                MF-TAPNet \cite{jin2019incorporating} & ~ & 67.92 & - \\ 
                \midrule 
                Duel-MF \cite{zhao2020learning} & \multirow{1}{*}{Semi.} & 62.51 & - \\
                \midrule 
                Ours (CLU) & Unsup. & 59.23 & 57.52 \\ 
                \bottomrule 
                
            \end{tabular}
            }
        \end{table}

        \begin{table}[htbp]
        \caption{
        Type Segmentation Performance \textmd{(mIoU [\%])}.
        }
        \label{Tab: TypesComparison}
        \centering
        \resizebox{1.0\linewidth}{!}{
            \begin{tabular}{l c c c c c}
        
                \toprule 
                ~ & Supervision & EndoVis2017 & EndoVis2018 \\ 
                \midrule 
                \midrule 
                BAANet \cite{shen2023branch} & \multirow{7}{*}{Sup.} & 61.59 & 42.67 \\ 
                DMNet \cite{wang2021efficient} & ~ & 53.89 & - \\
                SurgNet \cite{ni2022surginet} & ~ & 66.30 & - \\ 
                LWANet \cite{ni2020attention} & ~ & 58.30 & - \\ 
                ISINet \cite{gonzalez2020isinet} & ~ & 38.08 & 45.29 \\ 
                MF-TAPNet \cite{jin2019incorporating} & ~ & 36.62 & - \\ 
                SurgicalSAM \cite{Yue_Zhang_Hu_Xia_Luo_Wang_2024} & ~ & \underline{67.03} & \underline{58.87} & \\
                \midrule 
                Duel-MF \cite{zhao2020learning} & \multirow{1}{*}{Semi.} & 52.80 & -  \\
                \midrule 
                Ours (CLU) & Unsup. & 58.86 & 44.68 \\
                \bottomrule 
                
            \end{tabular}
            }
        \end{table}

        \begin{table}[htbp]
        \caption{
        Semantic Segmentation Performance \textmd{(mIoU [\%])}.
        }
        \label{Tab: SemanticComparison}
        \centering
        \resizebox{1.0\linewidth}{!}{
            \begin{tabular}{l c c c c c}
        
                \toprule 
                ~ & Supervision & EndoVis2018 & CholecSeg8k \\ 
                \midrule 
                \midrule 
                SwinSP-TCN \cite{grammatikopoulou2024spatio} & \multirow{2}{*}{Sup.} & - & \underline{69.38} \\ 
                Noisy-LSTM \cite{wang2021noisy} & ~ & \underline{62.30} & - \\ 
                \midrule 
                Ours (CLU) & Unsup. & 46.46 & 46.31 \\ 
                \bottomrule 
                
            \end{tabular}
            }
        \end{table}

        \begin{figure*}[htbp]
        \centering
        
        \includegraphics[width=0.9\linewidth]{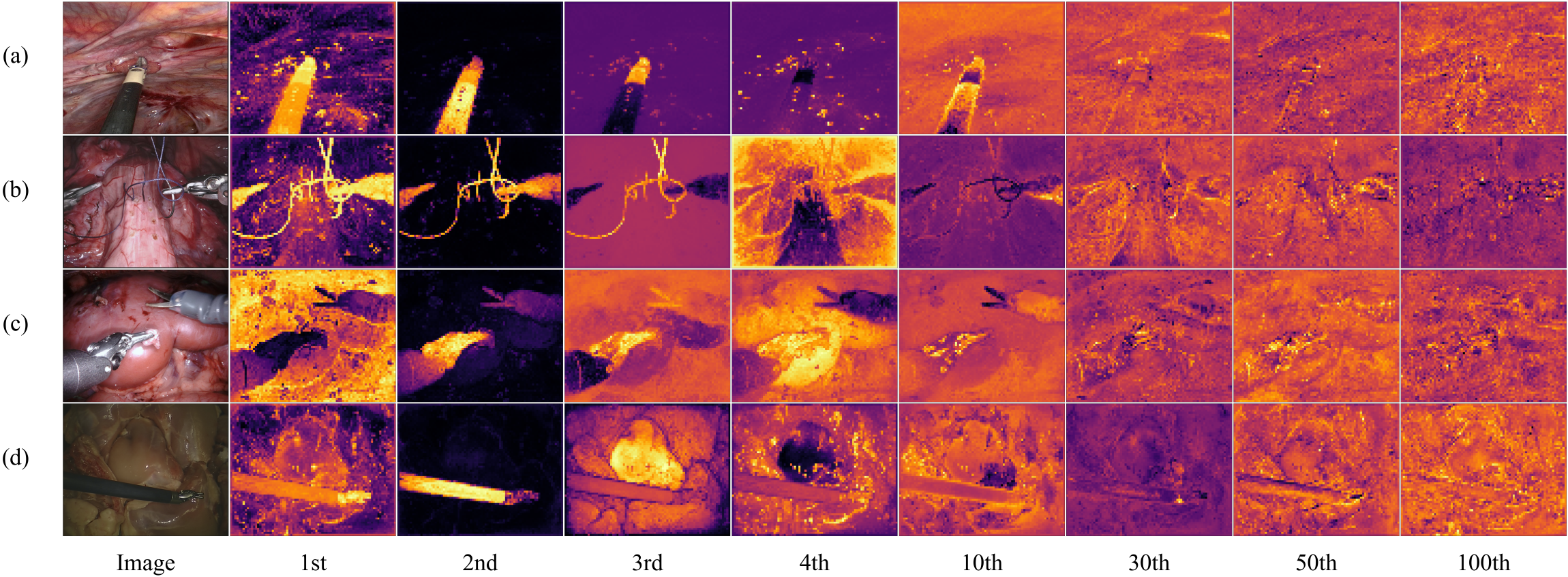}
        
        \caption{Eigenvectors Visualization. \textmd{
            Each row demonstrates the input image and its corresponding eigenvectors. (a) - (d) are from ARTNetDataset, EndoVis2017, EndoVis2018, and UCL respectively. "$i$ th" indicates the eigenvector with the $i$-th smallest eigenvalue. The Fiedler vector is denoted by "2nd".
        }}
        \label{Fig: EigenVis}
        \Description{}
        \end{figure*}

        \begin{figure}[htbp]  
        \centering
        
        \includegraphics[width=0.75\linewidth]{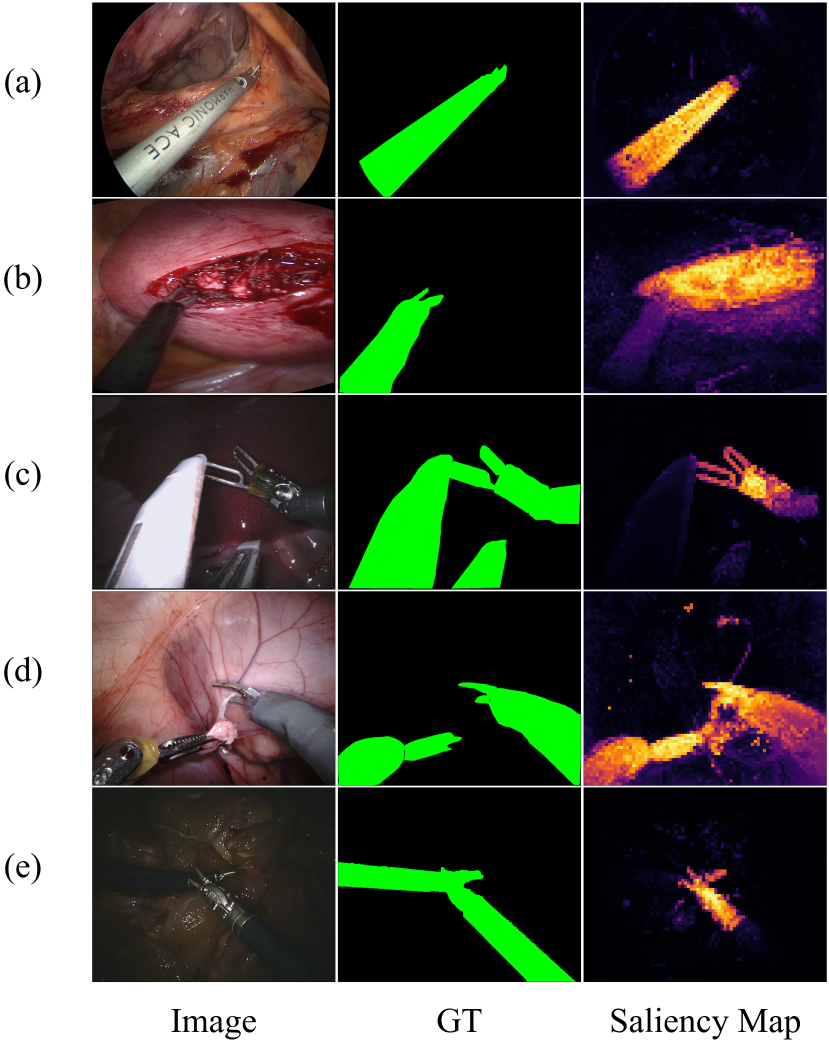}
        
        \caption{Salient Detection Visualization. \textmd{
            Each row illustrates an input image (Image), its binary ground-truth (GT), and the corresponding saliency map generated from the Fiedler vector. (a) and (b) are from ARTNetDataset; (c), (d), and (e) are from EndoVis2017, EndoVis2018 and UCL, respectively.
        }}
        \label{Fig: SalientDetection}
        \Description{}
        \end{figure}
        
        In Table \ref{Tab: PartsComparison}, we compare our method with supervised and semi-supervised methods on the part segmentation task. For the EndoVis2017 dataset, our method performance is close to the semi-supervised method Duel-MF with a small gap of 3\%, whereas there is still a large gap of 17\% compared to the fully supervised method DRLIS. Our method achieves a mIoU score of 57.52\% on the EndoVis2018 dataset, which is slightly lower than that of EndoVis2017 due to the difficulty and complexity of the EndoVis2018 dataset. The experimental results on the type segmentation task are reported in Table \ref{Tab: TypesComparison}, where our method shows medium-level performance at 58.86\% on EndoVis2017. However, relatively low performance is exposed on the EndoVis2018 dataset at 44.68\%, close to some supervised methods (e.g., BAANet and ISINet) but remaining a distinct gap (about 15\%) to the SOTA supervised method SurgicalSAM at 58.87\%. The experiment for the semantic segmentation task uses the EndoVis2018 and CholecSeg8k datasets, as demonstrated in Table \ref{Tab: SemanticComparison}. Although, we are the first unsupervised method dealing with the semantic segmentation task where the mIoU scores are 46.46\% (for EndoVis2018) and 46.31\% (for CholecSeg8k), supervised methods outperform our unsupervised method by 15.84\% and 23.07\% on EndoVis2018 and CholecSeg8k respectively.

        \subsection{Results Visualization}
        \label{ResultsVisualization}

        \begin{figure*}[htbp]
        \centering
        
        \includegraphics[width=0.9\linewidth]{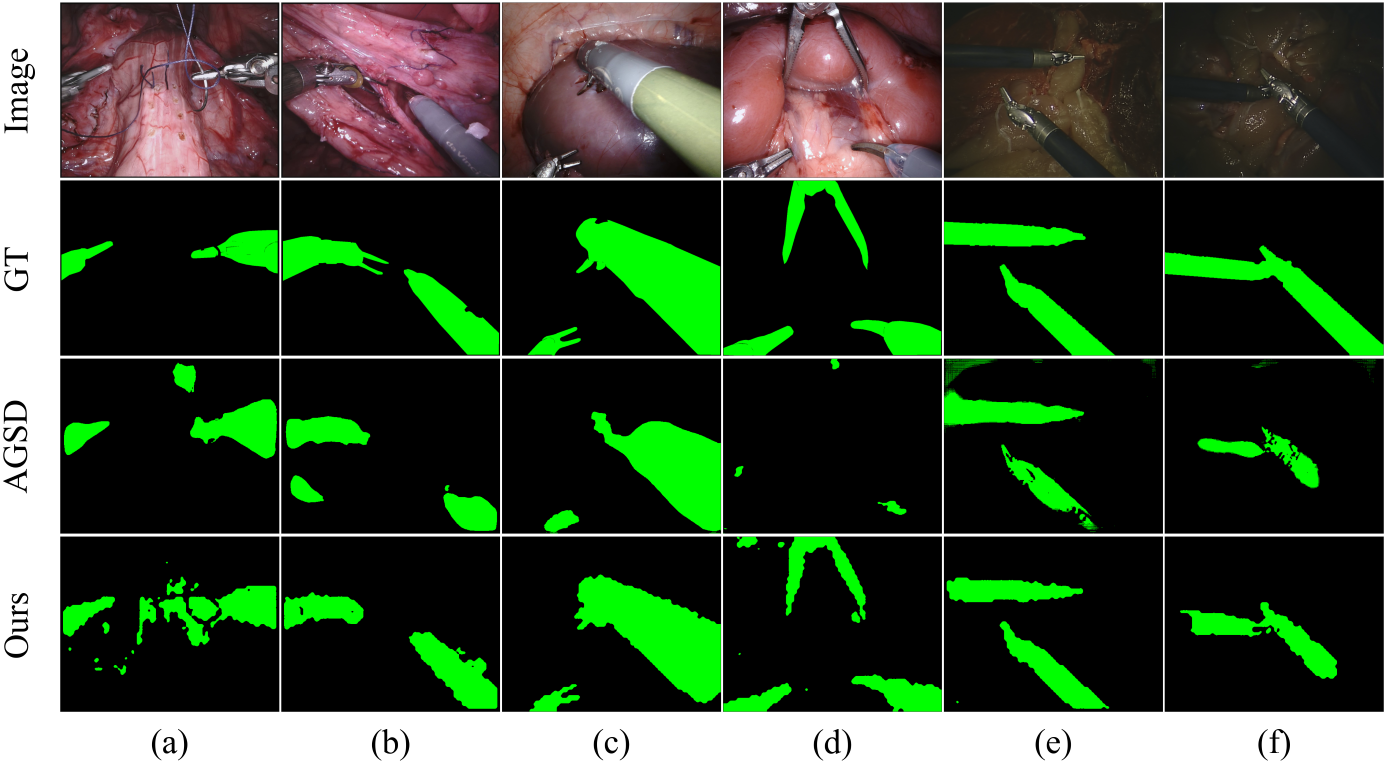}
        
        \caption{Binary Segmentation Visualization. \textmd{
            Each column demonstrates the input image (Image), binary ground-truth (GT), and prediction masks of AGSD and our CLU method. (a) and (b) are from EndoVis2017, (c) and (d) are from EndoVis2018, (e) and (f) are from UCL, where (a) shows a fail case of our CLU method.
        }}
        \label{Fig: BinaryComparison}
        \Description{}
        \end{figure*}
        

        \noindent
        \textbf{Eigenvectors Visualization.} Samples of eigenvectors are visualized in Figure \ref{Fig: EigenVis}, where every eigenvector reflects a particular module in its corresponding image. For example, in Figure \hyperref[Fig: EigenVis]{\ref{Fig: EigenVis}(a)}, the surgical tool is entirely detected in the first and second eigenvectors, suggesting sufficient information for the binary segmentation task, and in the third eigenvector, the part of the tool head is detected with the highest significance while the lowest magnitude for the tool shaft, thereby the part segmentation task can be easily solved. In terms of Figure \hyperref[Fig: EigenVis]{\ref{Fig: EigenVis}(b)}, combining the second and third eigenvector, the surgical tools and suturing thread are practically distinguished in the purpose of the type segmentation task. For the challenging semantic segmentation task, in Figures \hyperref[Fig: EigenVis]{\ref{Fig: EigenVis}(c) and \ref{Fig: EigenVis}(d)}, the organs and tissues are detected in the fourth and third eigenvectors respectively. The detection becomes more fine-grained, as eigenvalues increase, such as the "10th" eigenvector where more details like textures are detected. However, for the eigenvectors with overlarge eigenvalues (e.g., "30th", "50th" and "100th"), the detection becomes unintuitive and noise-like, which may negatively influence the method performance.
    
        
        \noindent
        \textbf{Salient Detection Visualization}. Samples of salient detection results on the binary SIS task are visualized in Figure \ref{Fig: SalientDetection}. The outstanding effectiveness of salient detection is illustrated in Figure \hyperref[Fig: SalientDetection]{\ref{Fig: SalientDetection}(a)} where the tool is detected as a salient object with high significance in the saliency map normalized from the corresponding Fiedler vector. However, the limitations and disadvantages of salient detection are displayed in Figures \hyperref[Fig: SalientDetection]{\ref{Fig: SalientDetection}(b) - \ref{Fig: SalientDetection}(e)}. The most common shortcoming is partial detection, such as in Figures \hyperref[Fig: SalientDetection]{\ref{Fig: SalientDetection}(c) - \ref{Fig: SalientDetection}(e)} where the instrument heads are detected as salient objects, whereas shafts are omitted or assigned low significance. Neglecting detection is another drawback as shown in Figures \hyperref[Fig: SalientDetection]{\ref{Fig: SalientDetection}(c) and \ref{Fig: SalientDetection}(e)} where only one instrument is partially detected, even though two or three tools appearing in the images. Another infrequent but severe disadvantage is misdiagnosis as shown in Figure \hyperref[Fig: SalientDetection]{\ref{Fig: SalientDetection}(b)} where the tool is barely detected with very low significance, while the surgical incision is emphasized with high significance.


        \noindent
        \textbf{Binary Segmentation Visualization.} Qualitative visualizations of our CLU method are illustrated in Figure \ref{Fig: BinaryComparison}. Our method is more capable of detecting the entire surgical tools, such as in Figures \hyperref[Fig: BinaryComparison]{\ref{Fig: BinaryComparison}(b) and \ref{Fig: BinaryComparison}(e)}. More fine-grained objects are practically segmented via our method, such as tool clamps in Figures \hyperref[Fig: BinaryComparison]{\ref{Fig: BinaryComparison}(c) and \ref{Fig: BinaryComparison}(d)}. For Figures \hyperref[Fig: BinaryComparison]{\ref{Fig: BinaryComparison}(b), \ref{Fig: BinaryComparison}(d) and \ref{Fig: BinaryComparison}(e)}, comparatively strong robustness of our method on lightless and over-lighted frames is demonstrated, while the severe light-reflection and the extreme darkness in the image frames significantly impact the AGSD method. However, some drawbacks are exposed in our method. For instance, the threads in Figure \hyperref[Fig: BinaryComparison]{\ref{Fig: BinaryComparison}(a)} are wrongly detected when they are connected with tools; and minor mis-segmentation on the light-reflection in Figure \hyperref[Fig: BinaryComparison]{\ref{Fig: BinaryComparison}(d)} that have a similar color to surgical instruments. Rough detection edges and low segmentation quality/resolution are inevitable disadvantages because the final results are obtained by interpolating the low-resolution prediction mask.

    \subsection{Ablation Studies}
    \label{ExpDis: AblationStudy}
        We conduct ablation experiments on the EndoVis2018 dataset because it supports all four segmentation tasks. As shown in Table \ref{Tab: AblationStudy}, the hyper-parameter $g$ influences performance promotion and degeneration. Increasing $k$ always leads to higher performance because of over-clustering. For few-class segmentation tasks (i.e., binary and part), the suitable $g$ is approximately located in fixed numbers such as $g=10$ for "Binary" and $g=15$ for "Part". In terms of complex segmentation tasks (i.e., type and semantic), the optimal $g$ depends on the number of clusters $k$, and $g=k$ often yields superior performance, as shown in Table \ref{Tab: AblationStudy} for "Type" and "Semantic". The main reason for the performance degeneration caused by overlarge $g$ is owed to the unintuitive and noise-like eigenvectors possessing large eigenvalues, as illustrated in Figure \ref{Fig: EigenVis}.
        
        \begin{table}[htbp]
        \caption{
        Ablation Study for Different $g$ and $k$.
        \textmd{"$g$" is the number of used eigenvectors. "$k$" is the number of clusters. \textbf{Bold} indicates the best result for each $k$.}
        }
        \label{Tab: AblationStudy}
        \centering
        \resizebox{1.0\linewidth}{!}{
            \begin{tabular}{c||c| c c c c c c c} 
                \toprule 
                ~ & \diagbox{$k$}{mIoU (\%)}{$g$} & 2 & 3 & 5 & 10 & 15 & 20 & 30 \\ 
                \midrule 
                \midrule 
                \multirow{5}{*}{Binary} & 5 & 73.54 & \textbf{76.05} & 73.80 & 69.28 & 65.14 & 64.02 & 61.75 \\ 
                ~ & 10 & 74.07 & 78.05 & 78.67 & \textbf{79.77} & 77.39 & 75.48 & 72.51 \\ 
                ~ & 15 & 74.17 & 78.53 & 79.71 & \textbf{80.20} & 79.46 & 78.81 & 77.07 \\ 
                ~ & 20 & 74.69 & 78.87 & 80.79 & \textbf{80.85} & 80.36 & 79.91 & 79.33 \\ 
                ~ & 30 & 75.01 & 79.21 & 81.82 & \textbf{82.24} & 81.82 & 81.46 & 81.20 \\ 
                \midrule 
                \midrule 
                \multirow{5}{*}{Part} & 5 & 44.69 & \textbf{48.77} & 48.58 & 45.27 & 41.54 & 40.13 & 37.80 \\ 
                ~ & 10 & 45.61 & 51.28 & 53.28 & \textbf{56.79} & 54.61 & 51.85 & 48.85 \\ 
                ~ & 15 & 45.82 & 51.68 & 54.13 & 57.13 & \textbf{57.52} & 56.85 & 55.10 \\ 
                ~ & 20 & 46.27 & 51.85 & 55.23 & 58.41 & \textbf{58.75} & 58.60 & 57.83 \\ 
                ~ & 30 & 46.52 & 52.11 & 56.24 & 59.50 & \textbf{59.94} & 59.86 & 59.79 \\ 
                \midrule 
                \midrule 
                \multirow{5}{*}{Type} & 5 & 30.49 & 32.81 & \textbf{34.35} & 31.54 & 28.64 & 26.70 & 25.06 \\ 
                ~ & 10 & 32.55 & 36.24 & 39.46 & \textbf{42.64} & 40.41 & 39.81 & 38.12 \\ 
                ~ & 15 & 33.69 & 36.81 & 40.81 & 43.49 & \textbf{44.68} & 43.99 & 42.36 \\ 
                ~ & 20 & 34.15 & 37.58 & 41.62 & 43.93 & 46.03 & \textbf{46.80} & 45.48 \\ 
                ~ & 30 & 34.99 & 38.01 & 42.34 & 44.70 & 47.51 & 48.29 & \textbf{49.70} \\  
                \midrule 
                \midrule 
                \multirow{5}{*}{Semantic} & 5 & 27.45 & 30.87 & \textbf{34.19} & 31.91 & 29.28 & 27.83 & 26.22 \\ 
                ~ & 10 & 29.85 & 34.61 & 39.56 & \textbf{43.06} & 42.14 & 41.31 & 39.72 \\ 
                ~ & 15 & 30.90 & 36.09 & 41.45 & 45.54 & \textbf{46.46} & 46.45 & 45.70 \\ 
                ~ & 20 & 31.57 & 36.94 & 42.85 & 47.15 & 48.57 & 49.16 & \textbf{49.70} \\ 
                ~ & 30 & 32.43 & 37.74 & 44.28 & 49.10 & 51.36 & 51.99 & \textbf{53.35} \\ 
                \bottomrule 
                \bottomrule 
            \end{tabular}
            }
        \end{table}

\section{Conclusion and Future Work}
\label{Conclusion}
    This work proposes an unsupervised SIS method based on graph theory, where a ViT-based model is used as a backbone to capture the global context information, and the image is segmented in a graph partitioning manner by regarding the pixels in the feature map as nodes in a graph. Then, the Laplacian matrix is calculated from the feature map and decomposed into "deep" eigenvectors representing particular semantic modules in an image (e.g., tools, threads, and organs), providing distinguishable information. The performance and robustness of our method are verified for the different SIS tasks (i.e., binary, part, type, and semantic) by conducting comprehensive experiments on various datasets such as CholedSeg8k, UCL, and EndoVis datasets.
    Nevertheless, due to the time-consuming eigendecomposition, low-resolution feature map, and unascertainable hyper-parameter $g$, computational inefficiency, relatively poor prediction quality, and extra effort for locating optimal $g$ are inevitable limitations of our methods.
    Therefore, future work should focus on developing a real-time, high-resolution, and adaptive unsupervised method that can generate high-quality prediction masks promptly and adaptively determine the hyper-parameter $g$, which may involve extra neural network structures, and design a strategy for locating the optimal $g$ that critically influences the method performance.
    
    




\bibliographystyle{ACM-Reference-Format}
\bibliography{main}

\end{document}